\documentclass[lettersize,journal]{IEEEtran}
\usepackage{amsmath,amsfonts}
\usepackage{array}
\usepackage{cite}
\usepackage{textcomp}

\usepackage{caption}
\usepackage{textcomp}
\usepackage[table,xcdraw]{xcolor}
\usepackage{ragged2e}
\usepackage{bbm}
\usepackage{multirow}
\usepackage{diagbox}
\usepackage[linesnumbered,ruled,vlined]{algorithm2e}
\usepackage[caption=false, font=footnotesize]{subfig}
\usepackage{mathtools} 
\usepackage[hang,flushmargin]{footmisc}
\usepackage{hyperref}
\hypersetup{colorlinks,urlcolor=magenta,linkcolor=blue,citecolor=blue}

\begin{document}

\title{CIFF-Net: Contextual Image Feature Fusion for Melanoma Diagnosis}

\author{Md Awsafur Rahman\textsuperscript{\textsection}, \IEEEmembership{Student Member, IEEE}, Bishmoy Paul\textsuperscript{\textsection}, \IEEEmembership{Student Member, IEEE}, Tanvir Mahmud, \IEEEmembership{Student Member, IEEE}, and~Shaikh Anowarul Fattah, \IEEEmembership{Senior~Member, IEEE}
\thanks{M.~A.~Rahman, B.~Paul, T.~Mahmud, and S.~A.~Fattah are with the Department of Electrical and Electronic Engineering, Bangladesh University of Engineering and Technology, Dhaka-1000, Bangladesh, e-mail: (awsaf49@gmail.com, paul.bish98@gmail.com, tanvirmahmud@eee.buet.ac.bd, and fattah@eee.buet.ac.bd).}}

\maketitle
\begingroup\renewcommand\thefootnote{\textsection}
\footnotetext{Equal contribution}
\endgroup

% The paper headers
\markboth{Journal of \LaTeX\ Class Files,~Vol.~XX, No.~XX, Febuary~2023}%
{MA Rahman \MakeLowercase{\textit{et al.}}: A Sample Article Using IEEEtran.cls for IEEE Journals}

\maketitle

\begin{abstract} 
Melanoma is considered to be the deadliest variant of skin cancer causing around 75\% of total skin cancer deaths. To diagnose Melanoma, clinicians assess and compare multiple skin lesions of the same patient concurrently to gather contextual information regarding the patterns, and abnormality of the skin. So far this concurrent multi-image comparative method has not been explored by existing deep learning-based schemes. In this paper, based on contextual image feature fusion (CIFF), a deep neural network (CIFF-Net) is proposed, which integrates patient-level contextual information into the traditional approaches for improved Melanoma diagnosis by concurrent multi-image comparative method. The proposed multi-kernel self attention (MKSA) module offers better generalization of the extracted features by introducing multi-kernel operations in the self attention mechanisms. To utilize both self attention and contextual feature-wise attention, an attention guided module named contextual feature fusion (CFF) is proposed that integrates extracted features from different contextual images into a single feature vector. Finally, in comparative contextual feature fusion (CCFF) module, primary and contextual features are compared concurrently to generate comparative features. Significant improvement in performance has been achieved on the ISIC-2020 dataset over the traditional approaches that validate the effectiveness of the proposed contextual learning scheme.
\end{abstract}


\section{Introduction}
\label{sec:introduction}
%write about importance of the melanoma detection
\IEEEPARstart{S}{kin} cancer is the most common cancer globally, with Melanoma being the most deadly form responsible for an overwhelming majority of skin cancer deaths~\cite{melanoma}. It is a major public health problem, with an estimated 106,110 cases expected to be diagnosed in US in 2021~\cite{cancer21}. Although the mortality is significant when detected early, Melanoma survival exceeds 95\%~\cite{isic20}. Finding Melanoma at an early stage is crucial; early detection can vastly increase the chances of cure~\cite{bray2018global}.

Existing approaches in melanoma detection have achieved expert-level performance in controlled studies examining single images using deep learning algorithms but they lack incorporation of patient-level information~\cite{isic20, xingguang2021deep, adewunmi2021enhanced, lee2020cancernet, pacheco2021attention, hmmbased, endtoend}. Recent studies have demonstrated the ability of AI-driven decisions to match, if not outperform, clinicians' decisions in the diagnosis of individual skin lesion images in controlled reader studies.However, the reader study did not accurately reflect clinical scenarios where clinicians have access to examine all lesions on a patient~\cite{isic20}.

%% figure 
\begin{figure}[t]
    \centering
    \includegraphics[scale=0.40]{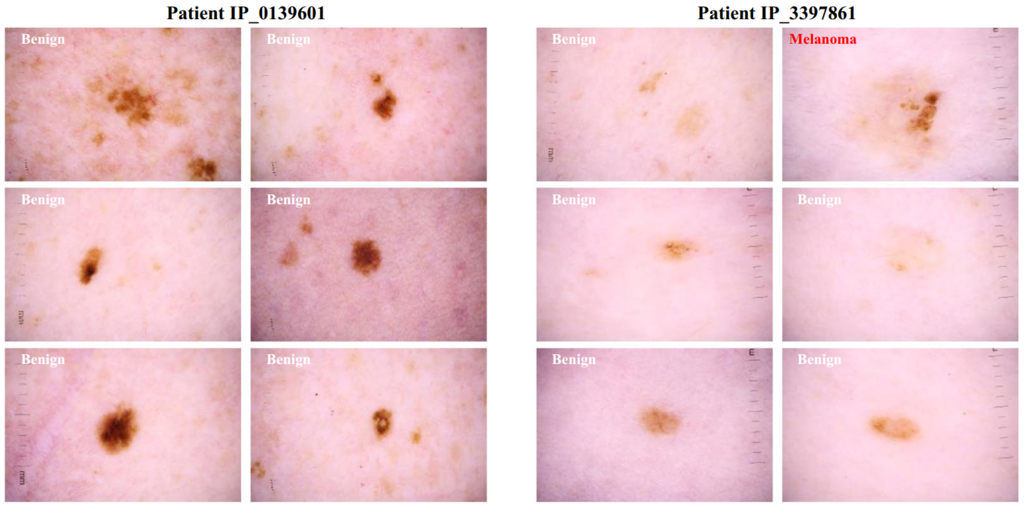}
    \caption{Example of clinical context. An atypical lesion found on a patient with many atypical lesions is less suspicious for malignancy as opposed to an atypical lesion that is an outlier on the patient.}
    \label{melanoma}
\end{figure}

\begin{figure*}[t]
    \centering
    \includegraphics[scale=0.15]{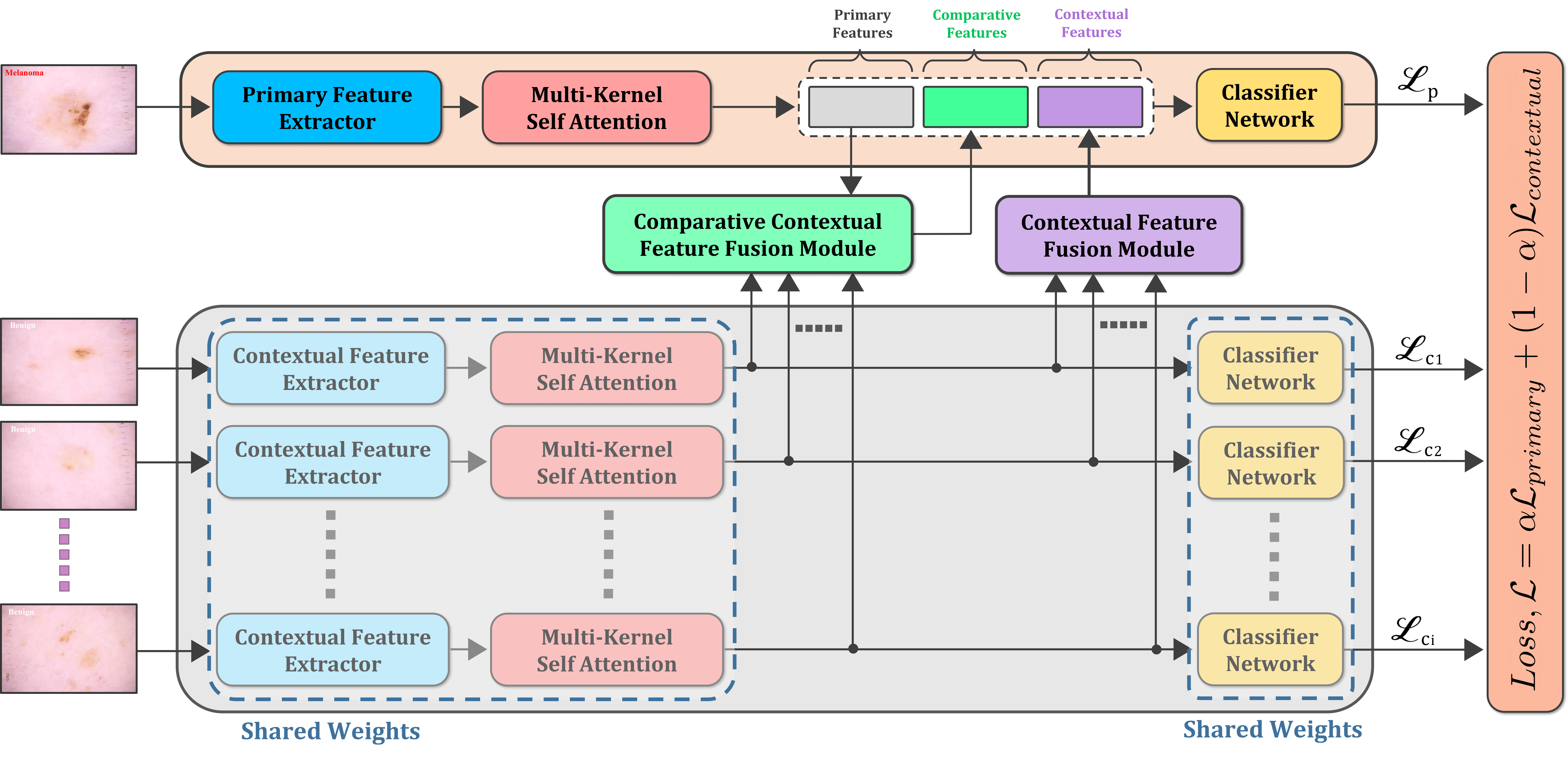}
    % \captionsetup{justification=raggedright}
    % \begin{margincap}
    \caption{An overview of proposed scheme. Primary image is passed to the Feature Extractor and MKSA module to produce Primary Feature vector. Whereas all contextual images are processed similarly, but with a different network, to produce intermediate features which are then used by the CFF and CCFF blocks to produce contextual and comparative features, respectively. All extracted features are used by Classifier to produce prediction with joint optimization using weighted loss.}
    % \end{margincap} 
    \label{abstract}
\end{figure*}

%how it is typically done by expert doctors
In practice, dermatologists base their judgment holistically on multiple lesions on the same patient~\cite{ud,isic20}. This patient-level contextual information is frequently used by clinicians to diagnose Melanoma and is especially useful in ruling out false positives in patients with many atypical nevi. Clinicians frequently assess skin lesions for biopsy by assessing them in context with the rest of the lesions on a given patient’s body
%, taking into consideration the individual biological skin characteristics
~\cite{isic20}. As demonstrated in Fig.~\ref{melanoma}, a lesion with malignancy-predictive features among many similar lesions is thought not to be as dangerous as an odd lesion on a patient whose other lesions are more benign-looking. The latter is known in dermatology as the ugly duckling sign and is frequently used to diagnose Melanoma ~\cite{ud, cvpr-ud}, especially in patients with multiple melanocytic lesions.

In this paper, a novel method is proposed (similar to clinicians' method for identifying Melanoma) to compare primary image with contextual images for Melanoma detection while addressing above mentioned issues of existing approaches. The major contributions of the proposed method are summarized below:

\begin{enumerate}
    \item Patient-level contextual information is incorporated which mimics expert clinicians' strategy to identify Melanoma images that have not been explored by existing approaches so far.
 
    \item To aid the feature extraction process of traditional deep learning models in Melanoma classification, a multi-kernel self attention module (MKSA) is proposed. This module employs attention on different scales to specifically enhance important features.
    
    \item To selectively concentrate on a few relevant contextual features while ignoring unnecessary ones, a contextual feature fusion (CFF) module is proposed. This module extracts necessary features from patient-level contextual information for primary image classification using an attention mechanism.

    \item To mimic the comparing strategy of clinicians, a comparative contextual feature fusion (CCFF) module is proposed. It compares primary and contextual images pairwise to generate comparative features for further processing.
    
    \item To optimize the proposed method for both primary and contextual images compound loss function was introduced. It takes ground truth and prediction for both primary and contextual images and outputs the cost for both primary and contextual images.
    
\end{enumerate} 

\section{Related Work}

\subsection{Hand-crafted methods}
At an earlier stage, many manually crafted features were proposed for skin lesion classification, including texture~\cite{stanley2007relative, ballerini2013color, ganster2001automated}. However, hand-crafted features are often not robust and have poor discriminative ability. They also require high computation due to high dimensions and struggle to capture deeper semantic information. As such, they are no longer competitive against CNN-based methods.

\subsection{CNN-based methods}
CNN-based approaches showed dominating performance in recent years. \cite{attique2021two} introduces a two-stream deep neural network information fusion framework with a fusion-based contrast enhancement module and a feature selection framework module for generating more discriminative features. \cite{mutual} proposed MB-DCNN model for simultaneous skin lesion segmentation and classification. Here different networks mutually transfer knowledge to facilitate each other in the process of accurately locating and classifying skin lesions. L-CNN~\cite{l-cnn} utilizes two feature extraction modules, a feature discrimination network, and model fusion strategy to improve recognition performance. In \cite{feature_agg}, the authors employed a methodology wherein deep features were extracted from multiple augmented versions of an image using ResNets. Subsequently, these features were aggregated using Fisher Vector (FV) encoding, and an SVM kernel was used to finally classify melanoma.

\subsection{Clinically inspired methods}
The Taxonomies model~\cite{taxonomies} introduces a hierarchical structure and an attention module to mimic the decision-making processes of doctors. However, it only uses one image as input, ignoring the practice of using multiple inputs, which is commonly used by dermatologists. Despite their successes, these methods are limited by their failure to incorporate medical expertise, thereby hindering the interpretability of their results. In~\cite{cvpr-ud} this problem is approached as outlier detection by variational autoencoder (VAE). Their approach utilizes reconstruction loss and distance of feature vectors to detect outliers. Even though their approach shows promising results, their model can only be trained on a single patient, and thus predicting for a new patient requires the model to be retrained on that patient's relevant data. More recently, the clinically inspired method (CI-Net)~\cite{ci-net} proposed a three-step process including zooming, observation, and comparison, along with a distinguishing strategy to simultaneously learn from two images and determine if they belong to the same class. However, this method only compares different features within a single image, while in practice clinicians compare multiple images to diagnose melanoma. Additionally, the distinguish strategy employed serves the purpose of contrastive learning rather than the "ugly duckling" method used by expert clinicians. Therefore, although it has achieved promising results, this method does not adequately incorporate patient-level contextual information.

\section{Methodology}
In Melanoma classification, additional lesion images of a patient can provide important context in providing a decision. For instance, in Fig.~\ref{melanoma}, even though all the lesions of patient \textit{IP\_0139601} seem severe, in truth they are benign as they all are similar. Whereas, in patient \textit{IP\_3397861}, even though the 2nd lesion sample(1st row, 2nd column) appears less severe, it is actually malignant as it is different from other samples. From Fig.~\ref{melanoma} it is evident that images with additional context can be a vital asset for improved diagnosis of Melanoma as opposed to the traditional method of inferring from a single image. 

The objective of the proposed scheme is to aid the conventional Melanoma classification process by incorporating information from other images from the same patient. In the experiment, the original image to classify is referred to as primary image and the images that provide additional context to the primary image classification are referred to as contextual images. An overview of proposed method is described in Fig.~\ref{abstract}.

\begin{figure}[t]
    \center
    \includegraphics[scale=0.12]{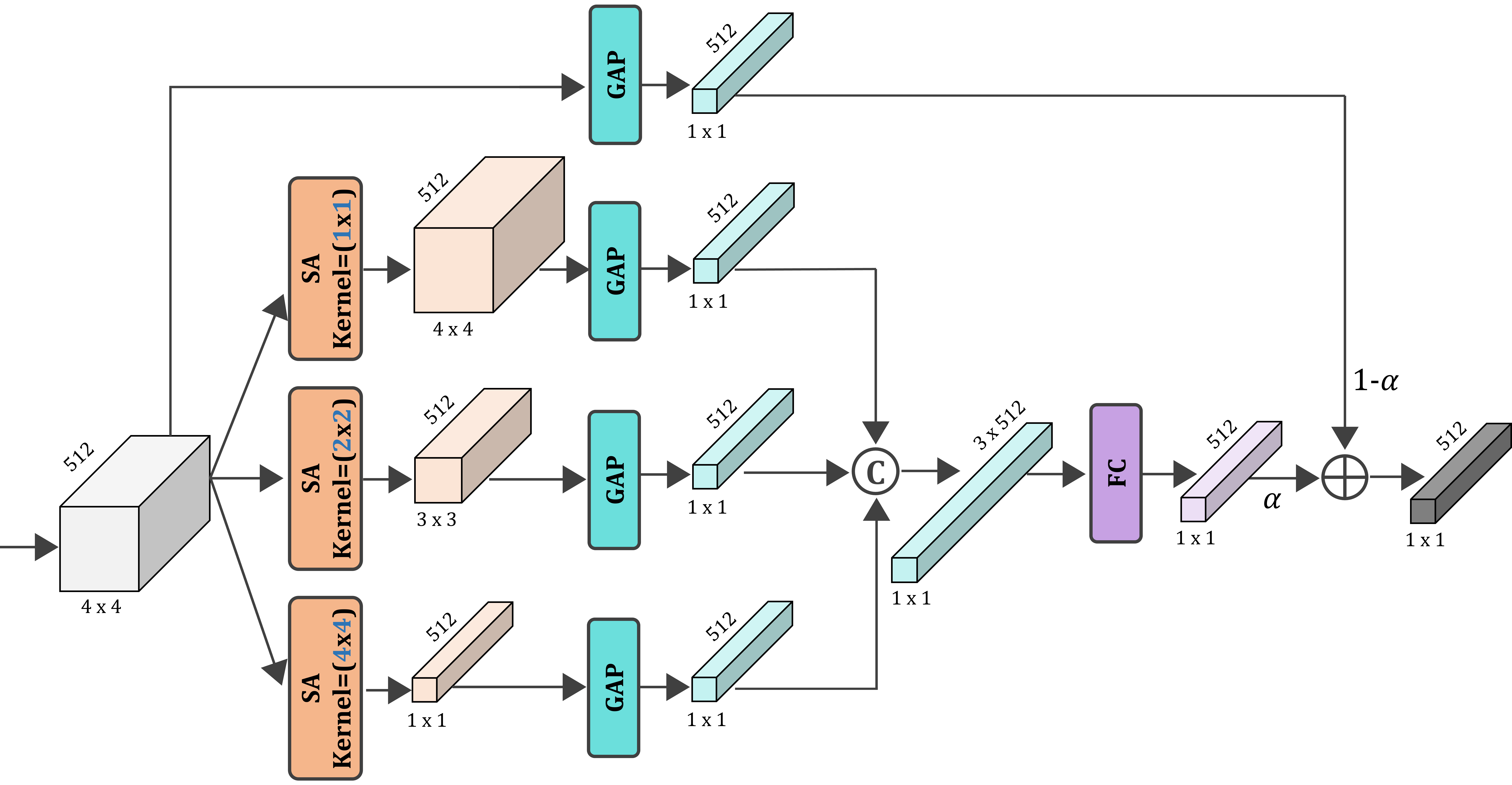}
    \caption{Multi-Kernel Self Attention (MKSA) Module. It consists of 3 parallel Self-Attention(SA) modules with different scales, whose outputs are merged and finally added to a Global Attention Pooled input. $\alpha$ is the learnable weight for addition.}
    \label{MKSA}
\end{figure}

\begin{figure}[t]
    \center
    \includegraphics[scale=0.12]{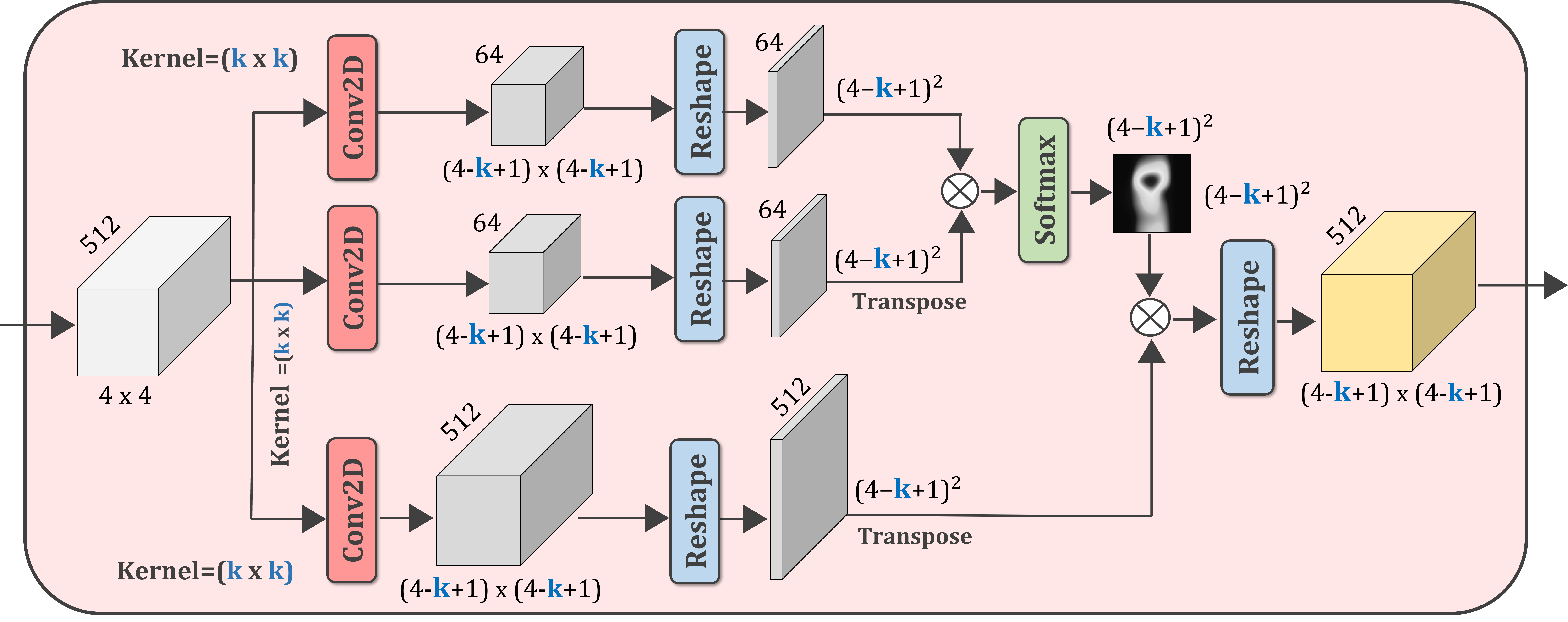}
    \caption{Self Attention(SA) Module. It has variable kernel shape, which allows for processing on multiple scales.}
    \label{SA}
\end{figure}

\subsection{Problem Formulation}

Let us consider the set of images and their corresponding ground truths are represented by $\mathbf{A}$, and $\mathbf{B}$, respectively, such that $A_i \in R^{h\times w \times c}$ and $B_i \in \{0, 1\}$. Here,  $i \in \{1,2,3, \dots N\}$ and $(h,w,c)$ denote height, width, and channel of a single image from the set of $N$ images.

To incorporate a set of contextual images on the diagnosis of a particular primary image, different sets of images can be formed where each set contains a primary image with several contextual images from a particular patient. Hence, a set of images, $X$, with $m$ contextual images,  and its corresponding ground truth, $Y$, can be represented as

\begin{align}
 &X_i = \{P_{i}, C_{1i}, C_{2i}, \dots, C_{mi}\}  \\
 &Y_i = \{y^p_i, y^c_{1i}, \dots, y^c_{mi}\} \\
 %&\mathbf{C_i} = \{ C_{1i}, C_{2i}, \dots, C_{mi} \}   \\
 &P, C \in \mathbf{A};\ y^p, y^c \in \mathbf{B}; X_i \in \mathbf{X}, Y_i \in \mathbf{Y} \nonumber \\
 &\forall i \in \{1, \dots, N\} \nonumber 
\end{align}

Here, $P_i$ denotes primary lesion of the $i_{th}$ set, and  $C_{ji}$ represents the $j_{th}$ contextual image of that particular set from the total $N$ lesions. Whereas, $y^p, \ y^c$ denote the ground truth of a primary and contextual lesion, respectively.

The objective is to extract optimum primary ($F_{primary}$), contextual feature ($F_{contextual}$), and comparative features ($F_{comparative}$) from each set of lesions $X$ for the precise diagnosis, and hence

\begin{align}
    &F_{primary}^i = \mathcal{F}(P_i) \\
    &F_{contextual}^i = \mathcal{F}(C_{1i}, \dots, C_{mi}) \\
    &F_{comparative}^i = \mathcal{F}(P_i, C_{1i}, \dots, C_{mi}) \\
    &\forall i \in \{1, \dots, N\} \nonumber
\end{align}

Hence, $F_{primary}$ represents only the primary lesion, $F_{contextual}$ considers all of the contextual lesions on a particular set, and $F_{comparative}$ considers each contextual lesion in comparison with the primary lesion of a particular set.

Afterward, the extracted contextual ($F_{contextual}$) and comparative features ($F_{comparative}$) are supposed to be integrated with the primary lesion feature ($F_{primary}$) for the diagnosis of the primary lesion, $y_{pred}^p$. Whereas, a set of diagnoses, ($y^c_{pred}$) is also provided for the contextual lesions that separately consider features from each contextual lesion. These can be represented as
\begin{align}
    & y^p_{pred, i} = \mathcal{FC}(F_{primary}^i, F_{contextual}^i, F_{comparative}^ i) \\
    & y^c_{pred, ij} = \mathcal{FC}(C_{ij});\ \forall j \in \{ 1, \dots, m\}\\
    & Y^{pred}_i = \{y^p_{pred, i}, y^c_{pred, 1i}, \dots, y^c_{pred, mi}\}\\
    & \forall i \in \{1, \dots, N\} \nonumber
\end{align}

Where $\mathcal{FC}$ denotes fully connected classifier layers. Finally, the objective function, $\mathcal{L}_{obj}$, can be defined as
\begin{align}
    & L_{obj} = \arg\min_{\theta} \mathcal{L}(\theta, \mathbf{X, Y, Y^{pred}})
\end{align}
where $\theta$ denotes the set of network parameters, $\mathbf{}$. Hence, the network parameter should be optimized for the precise diagnosis of the primary and contextual lesions.

\subsection{Proposed Multi Kernel Self Attention (MKSA) Module}
Conventional methods of feature extraction from images rely on CNN network as feature extractor to generate 2D feature maps which are flattened and pooled differently across the channels to generate the final feature vector. But this method fails to extract inter-area or inter-pixel relations between two non-adjacent areas in the feature maps. The proposed Multi-kernel Self Attention (MKSA) module shown in Fig.~\ref{MKSA} can generate potent feature vectors by examining inter-area relations between generated feature maps.

The MKSA module comprises parallel branches of Self Attentions (SA). The self attention kernels shown in Fig.~\ref{SA} apply an attention mask to the input feature map that prioritizes useful information. These attention maps are Global Average Pooled to produce 1D feature vectors. Then feature vectors are concatenated, and a fully connected layer is used to reduce their dimension. Then original vector is added to processed vector in a residual manner with learnable weight.

Conventional self attention blocks, such as introduced in \cite{NonLocal2018} only examine the inter-pixel relations whereas in the proposed form of self attention kernel(SA Kernel in Fig.~\ref{SA}), the inter-pixel relations as well as the inter-area relations. The MKSA module consists of parallel branches of SA block with varying kernel sizes (e.g: 1x1, 2x2, and 4x4 kernel size) to extract features on multiple field-of-vision.

In this experiment, conventional CNN networks such as VGG\cite{vgg16}, Resnet\cite{resnet} and Densenets\cite{densenet}, EfficientNet\cite{efficientnet}
are used and since EfficientNets yield the highest baseline scores, they are prioritized in subsequent ablation studies.

\begin{figure}[t]
    \center
    \includegraphics[scale=0.13]{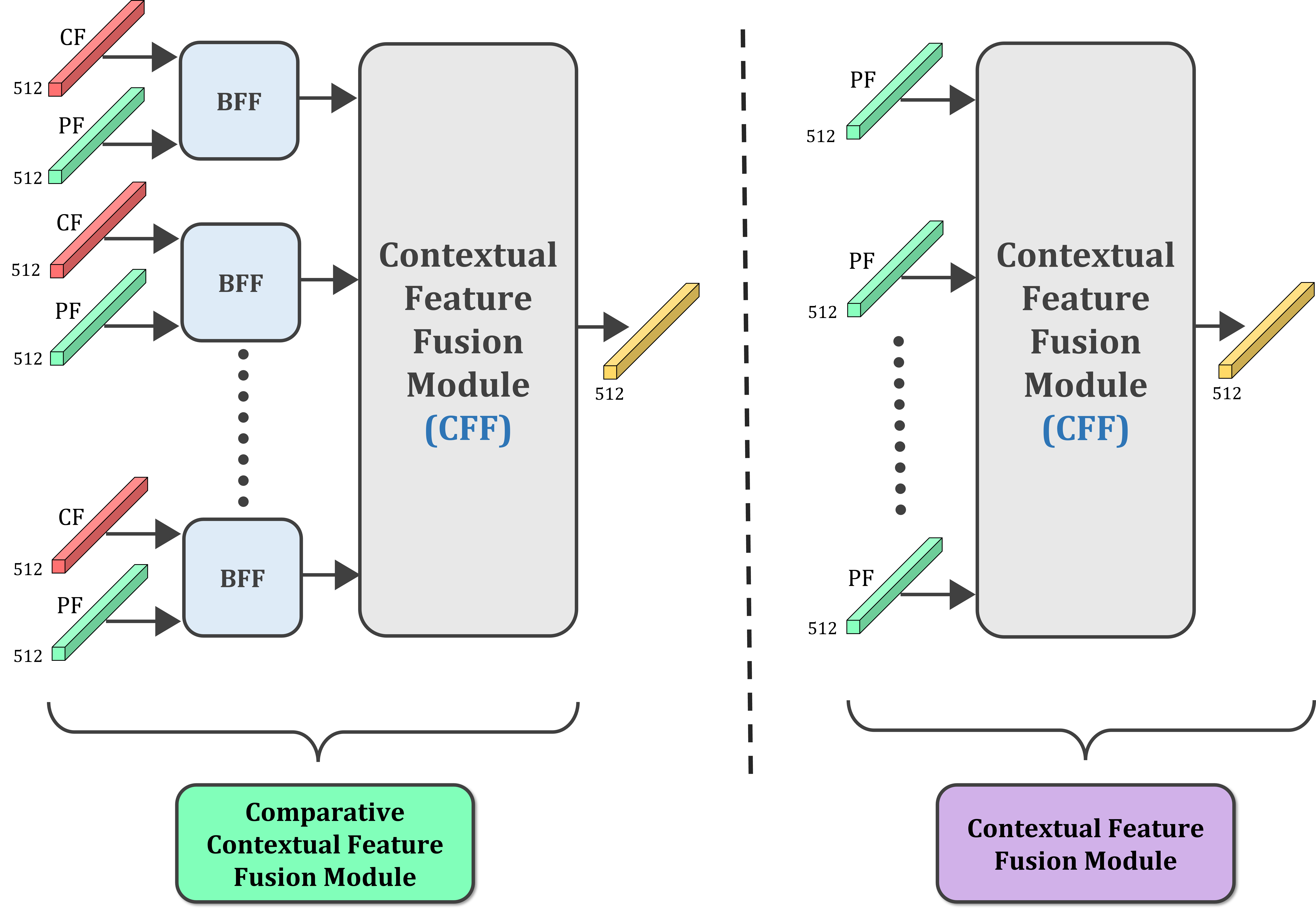}
    \caption{Comparative Contextual Feature Fusion (CCFF) Module (on the left) comprises of Contextual Feature Fusion (CFF) Module (on the right) and BFF Module.}
    \label{CCFF}
\end{figure}

\begin{figure}[t]
    \center
    \includegraphics[scale=0.12]{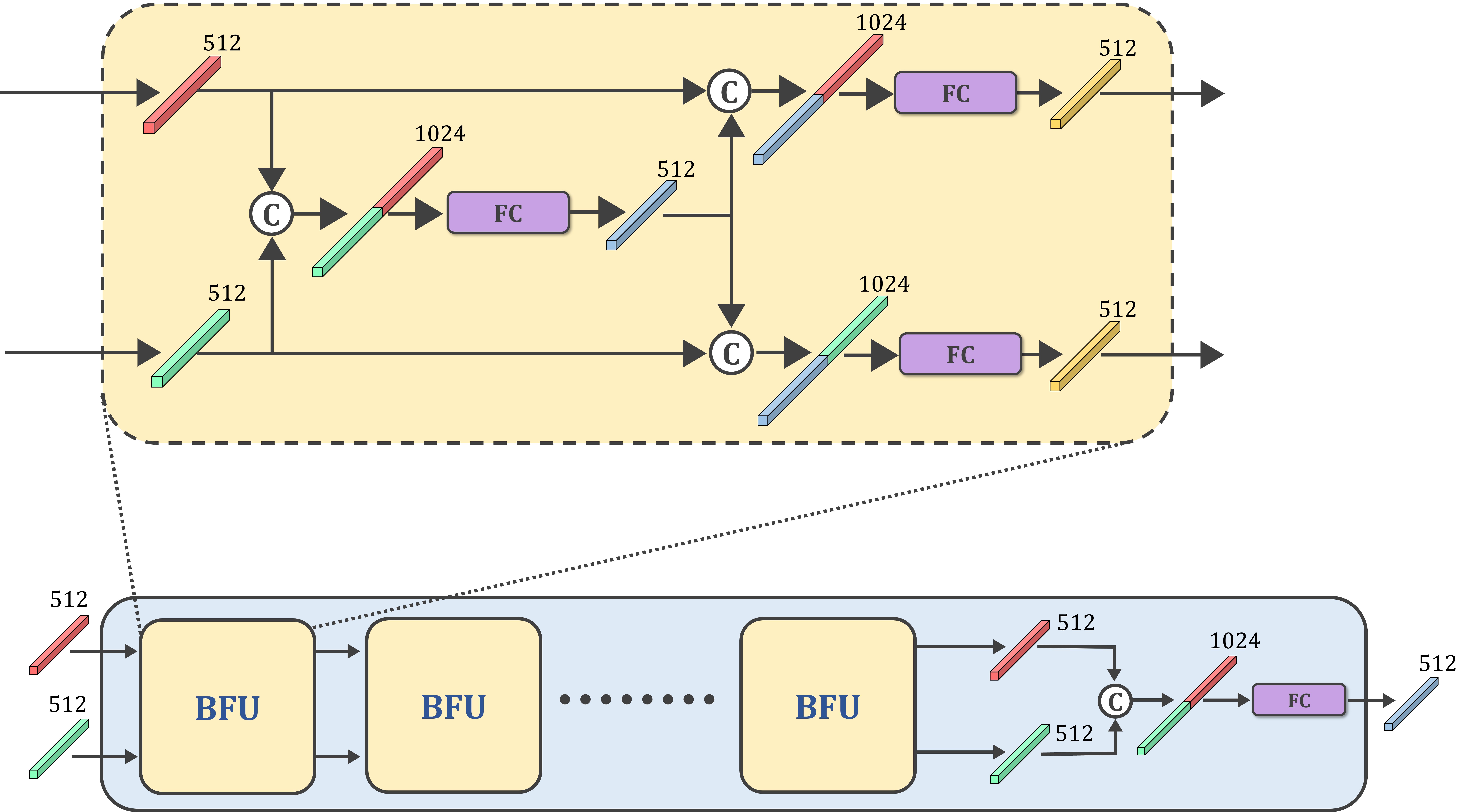}
    \caption{Binary Feature Fusion (BFF) Module comprises cascaded Binary Fusion Unit (BFU) Modules. The cascaded BFU modules allow for added depth to process the features.}
    \label{BFF}
\end{figure}

%\subsection{Proposed Attention Guided Feature Fusion (AGFF) Module}

\subsection{Proposed Contextual Feature Fusion (CFF) Module}

\begin{figure}[t]
    \center
    \includegraphics[scale=0.10]{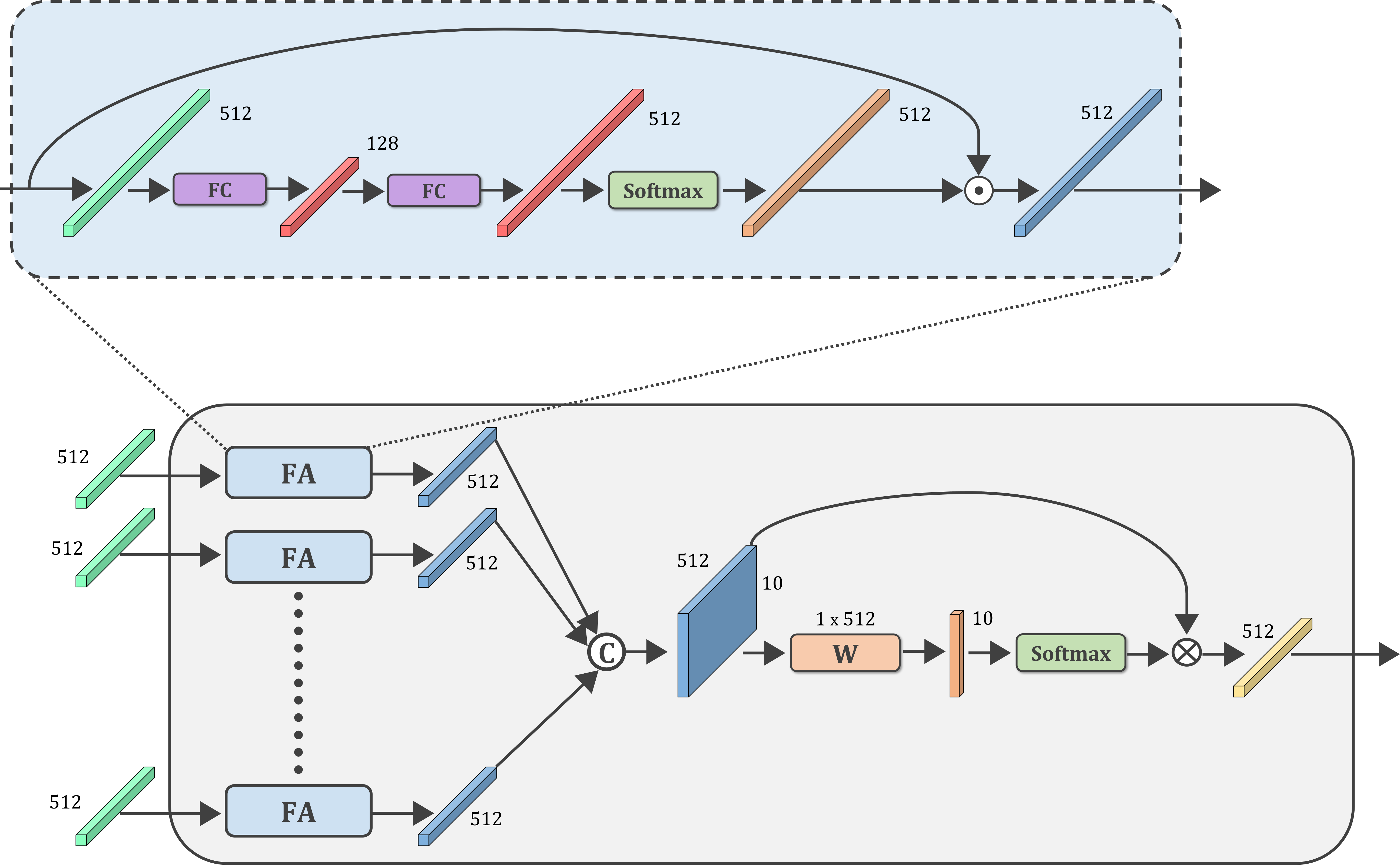}
    \caption{Contextual Feature Fusion(CFF) Module. It takes multiple feature vectors as input and generates a single feature vector as output.}
    \label{CFF}
\end{figure}

Contextual feature fusion (CFF) module shown in Fig.~\ref{CFF} is proposed to generate a contextual feature vector by passing the contextual image vectors to the CFF module. As input, the CFF module takes a group of feature vectors which are processed individually by a Feature Attention(FA) submodule. This submodule consists of dense layers that process the inter-feature relations of the input and assign a probability mask to them. This probability mask is applied to the input feature vector. So in the resultant output feature mask, the significant features are prioritized. The processed feature vectors generated from multiple contextual images are then concatenated.

The subsequent submodule focuses on assigning a probability mask in each of the group images based on the information available from the processed vectors and expressing them in a smaller vector. This submodule takes a weighted sum of each of the features in the processed vectors by multiplying the concatenated matrix with a weight vector with the same length as the feature vector. The softmax of this weighted sum assigns a probability of importance across the features in the feature vectors. The final output is a weighted sum of the same features of multiple image feature vectors.

\subsection{Proposed Comparative Contextual Feature Fusion (CCFF) Module}
Siamese Neural Networks\cite{siamese} introduced the idea of comparing image feature vectors generated from the same model to find similarities or dissimilarities between them. When the image feature vectors are generated from the same feature extractor model, the l1-distance between the two feature vectors can represent the feature similarities between the two images.

But utilizing this idea to suit the case of comparing the group images with the single image poses a problem:  the same feature extractor is not being used to generate feature vectors for the single image (primary image) which is being classified and the group of images that are providing contextual information (contextual images). Since utilizing the contextual feature vectors to aid the primary image classification and utilizing the primary feature vector to classify the primary image are separate tasks, the feature extractors assigned to them would create separate features. Finding any relation between two generated feature vectors would require additional processing of the feature vectors using information from both features. The Comparative and Contextual Feature Fusion(CCFF) module is proposed to examine the features to generate a representation of the similarity or dissimilarity between the two different types of feature vectors. The CCFF module is comprised of two submodules: Binary Feature Fusion(BFF) and the Contextual Feature Fusion Module(CFF) discussed above.

The Binary Feature Fusion(BFF) submodule (shown in Fig.~\ref{BFF}) takes two feature vectors and learns to compare them for extracting underlying relations. This is done by concatenating the two vectors and propagating them through a densely connected layer. This layer finally generates a comparative feature vector from these two feature vectors. 

To model more complex features with comparative information, a cascaded version of this submodule is proposed and the original submodule is extended to the module in Fig. ~\ref{CCFF}. In this extended submodule, the original comparative feature vector is concatenated with the two original inputs separately. The two resultant vectors are passed through densely connected layers respectively. So the comparative feature vector and the original input feature vector are jointly processed to create better inputs for the subsequent layers of the cascaded submodule and the generated features are deeper. At the final cascaded layer, the processed inputs generate a single comparative feature vector by the above-mentioned method of concatenation and subsequent propagation into a fully connected layer. Each unit of the cascaded submodule is named Binary Feature Unit(BFU). The experiments performed displayed that the cascaded submodule performs better in extracting deep features compared to a single BFU unit, so the cascaded submodule is used as part of the proposed solution.

For each of the contextual images present, its feature vector and the primary image feature vector are utilized as inputs in the proposed BFF submodule and the resultant feature vectors are passed into the proposed CFF submodule, which finally generates a single feature vector containing the information processed by the proposed submodules.

\begin{table*}
\centering
\caption{Comparison of Performances of Different Modules with Different Backbones}
\label{t1}
\resizebox{\textwidth}{!}{%
\begin{tabular}{|l|c|l|l|l|l|l|l|l|l|l|l|l|l|} 
\hline
\multicolumn{1}{|c|}{\multirow{2}{*}{Modules}}                                                  & \multirow{2}{*}{Contextual Image} & \multicolumn{4}{c|}{DenseNet121}                                                                                                                                                                              & \multicolumn{4}{c|}{EfficientNetB3}                                                                                                                                                                           & \multicolumn{4}{c|}{EfficientNetB5}                                                                                                                                                                            \\ 
\cline{3-14}
\multicolumn{1}{|c|}{}                                                                          &                          & \multicolumn{1}{c|}{Spec.} & \multicolumn{1}{c|}{Sens.} & \multicolumn{1}{c|}{\begin{tabular}[c]{@{}c@{}}AUC \\ (cv)\end{tabular}} & \multicolumn{1}{c|}{\begin{tabular}[c]{@{}c@{}}AUC \\ (lb)\end{tabular}} & \multicolumn{1}{c|}{Spec.} & \multicolumn{1}{c|}{Sens.} & \multicolumn{1}{c|}{\begin{tabular}[c]{@{}c@{}}AUC \\ (cv)\end{tabular}} & \multicolumn{1}{c|}{\begin{tabular}[c]{@{}c@{}}AUC \\ (lb)\end{tabular}} & \multicolumn{1}{c|}{Spec.} & \multicolumn{1}{c|}{Sens.} & \multicolumn{1}{c|}{\begin{tabular}[c]{@{}c@{}}AUC \\ (cv)\end{tabular}} & \multicolumn{1}{c|}{\begin{tabular}[c]{@{}c@{}}AUC \\ (lb)\end{tabular}}  \\ 
\hline
Baseline                                                                                        & No                       & 0.820                      & 0.192                      & 0.87                                                                     & 0.867                                                                    & 0.916                      & 0.234                      & 0.924                                                                    & 0.914                                                                    & 0.943                      & 0.254                      & 0.932                                                                    & 0.928                                                                     \\ 
\hline
\textbf{Baseline + MKSA}                                                                        & No                       & 0.827                      & 0.220                      & 0.882                                                                    & 0.874                                                                    & 0.920                      & 0.260                      & 0.925                                                                    & 0.916                                                                    & 0.947                      & 0.283                      & 0.938                                                                    & 0.925                                                                     \\ 
\hline
\textbf{Baseline + CFF}                                                                         & Yes                      & 0.830                      & 0.255                      & 0.891                                                                    & 0.883                                                                    & 0.933                      & 0.308                      & 0.925                                                                    & 0.917                                                                    & 0.951                      & 0.330                      & 0.941                                                                    & 0.930                                                                     \\ 
\hline
\begin{tabular}[c]{@{}l@{}}\textbf{Baseline + CCFF }\\\textbf{ (BFF w 1 BFU only)}\end{tabular} & Yes                      & 0.831                      & 0.267                      & 0.893                                                                    & 0.883                                                                    & 0.921                      & 0.310                      & 0.923                                                                    & 0.912                                                                    & 0.941                      & 0.339                      & 0.940                                                                    & 0.929                                                                     \\ 
\hline
\textbf{Baseline + CCFF}                                                                        & Yes                      & 0.833                      & 0.274                      & 0.898                                                                    & 0.887                                                                    & 0.928                      & 0.315                      & 0.926                                                                    & 0.918                                                                    & 0.956                      & 0.342                      & 0.944                                                                    & 0.933                                                                     \\ 
\hline
\begin{tabular}[c]{@{}l@{}}\textbf{Baseline + CFF +}\\\textbf{ CCFF}\end{tabular}               & Yes                      & 0.842                      & 0.311                      & 0.902                                                                    & 0.892                                                                    & 0.935                      & 0.352                      & 0.927                                                                    & 0.920                                                                    & 0.961                      & 0.361                      & 0.948                                                                    & 0.938                                                                     \\ 
\hline
\begin{tabular}[c]{@{}l@{}}\textbf{Baseline + MKSA +}\\\textbf{ CCF}\end{tabular}               & Yes                      & \multicolumn{1}{c|}{0.840} & \multicolumn{1}{c|}{0.323} & \multicolumn{1}{c|}{0.910}                                               & 0.895                                                                    & \multicolumn{1}{c|}{0.946} & \multicolumn{1}{c|}{0.367} & \multicolumn{1}{c|}{0.929}                                               & 0.924                                                                    & \multicolumn{1}{c|}{0.956} & \multicolumn{1}{c|}{0.379} & \multicolumn{1}{c|}{0.950}                                               & 0.940                                                                     \\ 
\hline
\begin{tabular}[c]{@{}l@{}}\textbf{Baseline + MKSA +}\\\textbf{ CCFF}\end{tabular}              & Yes                      & \multicolumn{1}{c|}{0.841} & \multicolumn{1}{c|}{0.342} & \multicolumn{1}{c|}{0.913}                                               & 0.899                                                                    & \multicolumn{1}{c|}{0.950} & \multicolumn{1}{c|}{0.380} & \multicolumn{1}{c|}{0.931}                                               & 0.925                                                                    & \multicolumn{1}{c|}{0.964} & \multicolumn{1}{c|}{0.395} & \multicolumn{1}{c|}{0.952}                                               & 0.942                                                                     \\ 
\hline
\begin{tabular}[c]{@{}l@{}}\textbf{Baseline + CFF + }\\\textbf{ MKSA + CCFF}\end{tabular}       & Yes                      & \multicolumn{1}{c|}{0.850} & \multicolumn{1}{c|}{0.350} & \multicolumn{1}{c|}{0.915}                                               & 0.907                                                                    & \multicolumn{1}{c|}{0.954} & \multicolumn{1}{c|}{0.390} & \multicolumn{1}{c|}{0.932}                                               & 0.927                                                                    & \multicolumn{1}{c|}{0.968} & \multicolumn{1}{c|}{0.401} & \multicolumn{1}{c|}{0.957}                                               & 0.945                                                                     \\
\hline
\end{tabular}}
\end{table*}

\begin{table}
\centering
\caption{Comparison with Different Existing Approaches}
\label{t10}
\scalebox{1.2}{
\begin{tabular}{|l|l|l|l|l|} 
\hline
\diagbox{\textbf{Method}}{\textbf{Score}} & \textbf{AUC}   & \textbf{Acc.}  & \textbf{Sens.} & \textbf{Spec.}  \\ 
\hline
\textbf{Proposed}                         & \textbf{0.957} & \textbf{0.983} & \textbf{0.401} & \textbf{0.968}  \\ 
\hline
CI-Net~\cite{ci-net}                                & 0.920          & 0.955           & 0.321          & 0.914           \\ 
\hline
EfficientNet~\cite{efficientnet}                         & 0.911           & 0.942          & 0.203          & 0.980           \\ 
\hline
Taxonomies~\cite{taxonomies}                           & 0.907          & 0.950          & 0.282          & 0.931           \\ 
\hline
SE-Net~\cite{se-net}                                & 0.908          & 0.944          & 0.135          & 0.898           \\ 
\hline
L-CNN~\cite{l-cnn}                                & 0.904          & 0.951           & 0.273          & 0.913           \\ 
\hline
MB-DCNN~\cite{mutual}                              & 0.904            & 0.944          & 0.178          & 0.942           \\ 
\hline
CBAM~\cite{cbam}                                 & 0.896          & 0.938          & 0.153           & 0.926           \\ 
\hline
DenseNet~\cite{densenet}                             & 0.889          & 0.936          & 0.316          & 0.903           \\ 
\hline
ResNet~\cite{resnet}                            & 0.884           & 0.940          & 0.254          & 0.897           \\
\hline
\end{tabular}}
\end{table}

\subsection{Classifier Network}
For the final classification from feature vectors, fully connected layers are used. The primary classifier uses a non-linear combination of primary, contextual, and comparative features to classify while the contextual classifier only uses information from contextual features. They accumulate high-level convolutional features to find the best non-linear combination to classify images.

\subsection{Proposed Loss Function}
The primary optimization goal of the proposed model is to generate accurate predictions made by the primary classifier. However, allowing the contextual classifier to generate predictions for the group of images used as contextual information and optimizing those predictions too is a secondary task, since optimizing this would aid the ability of the contextual feature extractor and classifier to extract meaningful information from the group images.   

For a single group of $N$ images, where $(N-1)$ images are supplementary and a single image is primary, the losses are calculated as:

\begin{align}
& \mathcal{L}_{contextual} = \frac{1}{N-1}\sum_{i=1}^{N-1}\mathcal{L}(y_{context_{i}},\hat{y}_{context_{i}})\\
& \mathcal{L}_{primary} = \mathcal{L}(y_{primary},\hat{y}_{primary} )\\
& \mathcal{L}_{total}=\alpha*\mathcal{L}_{primary}+(1-\alpha)*\mathcal{L}_{contextual}
\end{align}

Here $\mathcal{L}$ is the loss function and $\alpha$ can be adjusted to selectively prioritize contextual feature quality and primary prediction result. For our experiments, we used $\alpha = 0.8$.
The Total Loss is finally summed over the batches and used to calculate gradients using backpropagation. For our experiments, we used Binary Cross-entropy as both the primary loss function and contextual loss function.

\subsection{Proposed Multi-Phase Optimization Algorithm}
For the experiments, different conventional image feature extraction models aided with the MKSA module are used as Feature Extractors. In the first phase, the Feature Extractor is trained conventionally as a binary classifier with all the images without any patient-level contextual information so it utilizes information from only the primary image. Additionally, the feature extractors are initialized with pre-trained Imagenet weights, which creates a better starting point for the optimization algorithm. After the optimization, the Feature Extractor becomes capable of generating feature maps from the primary image. However, as this phase lacks patient-level contextual information, this phase can't reach the optimal result.

In the second phase, the training is done with the scheme shown in Fig.~\ref{abstract}. This scheme utilizes all the contextual images along with primary images. To preserve the Feature Extractor's learning, its weights are kept frozen in this phase. The output of feature extractors and MKSA module is propagated to the CFF and CCFF modules, where they create potent features which aid in the classification of the primary image. A secondary task of predicting the contextual image is also included to aid the optimization process.    

In the third phase, to introduce further optimization all modules are fine-tuned keeping their weights trainable. In this phase, all modules learn simultaneously with one another which is essential for optimal performance. Hence, this scheme is capable of exploiting contextual patient-level information along with single-image information which results in optimal performance.

\section{Results and Discussions}

\subsection{Dataset Description}
%describe the datasets
We extensively conduct various experiments on ISIC 2020 Challenge Dataset~\cite{isic20}. This dataset is based on the larger ISIC archive that contains the largest publicly available collection of quality-controlled dermoscopic images of skin lesions generated by the international skin imaging collaboration (ISIC). All images from this dataset include some information corresponding to the diagnosis and are labeled by some medical research institutes. The dataset represents $2,056$ patients ($20.8$\% with at least one Melanoma) from three continents, consisting of $33,126$ dermoscopic images. Among those images, $584$ (1.8\%) images are of histopathologically confirmed Melanoma compared with benign Melanoma mimickers. Each patient has a number of images ranging from $2$ to $115$. Among all the patients, $1143$ patients have images more than $10$ having $237$ Melanoma cases. Notably, experiments are conducted on ISIC20~\cite{isic20} dataset due to the lack of alternative datasets with comparable contextual information, as far as our current knowledge extends

\subsection{Experimental Setup}
%tools
Different hyper-parameters of the proposed network are chosen through extensive experimentation for better performance. Adam optimizer is employed for the optimization of the network during the training of all three phases. In the first phase, the model is trained with a learning rate of $\gamma_1=10^{-3}$ along with ReduceLROnPlateau~\cite{reducelr} scheduler. In the second phase, the learning rate is reduced to $\gamma_2=10^{-4}$ for proper learning of CFF, CCFF, and MKSA modules and in the third phase, the learning rate is further reduced to $\gamma_3=10^{-5}$ for the preservation of feature extractor's weights and proper fine-tuning. For experimentation, $2\times$ NVIDIA RTX $3090$ GPU is used. The network is trained for $100$ epochs with batch size 32. For proper validation, a five-fold cross-validation scheme is carried out. To avoid data leakage data split is kept identical for all three phases. It is ensured that the images kept for the training stage do not use in the validation stage. To prevent early over-fitting, simple augmentations namely, Random-Flip, Random-Rotation, Random-Shift, and Random-Zoom are used.

\begin{figure*}[t]
    \centering
    \includegraphics[scale=0.050]{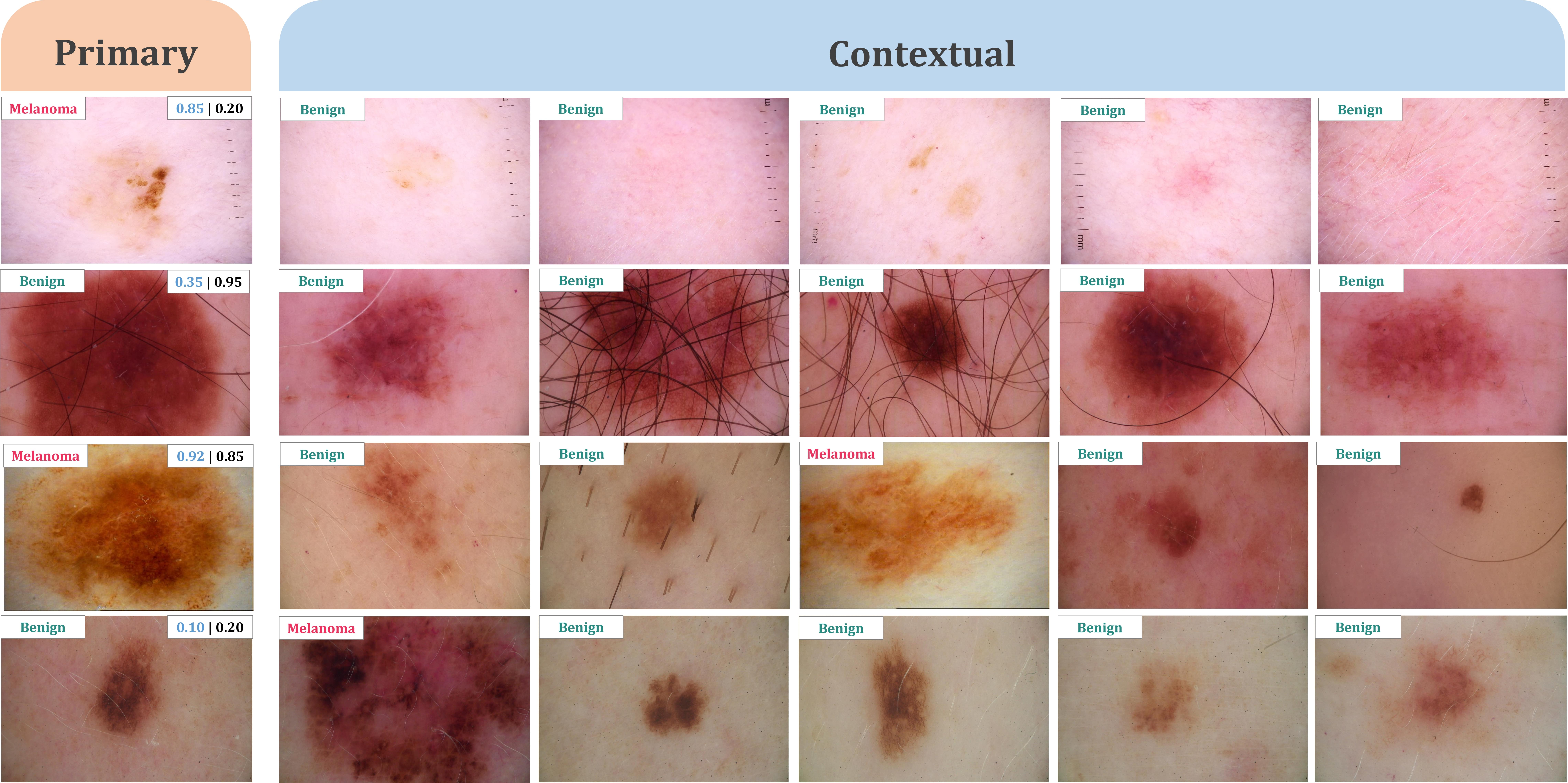}
    \caption{Visual comparison of the proposed method with a traditional single image-based method. Each primary image has ground truth (top left) and prediction (top right) displayed in blue and black, respectively, with and without contextual images}
    \label{pvsc}
\end{figure*}

\subsection{Comparison with existing SOTA approaches}
The proposed solution outperforms existing SOTA single model approaches on both natural image classification and skin lesion classification, as shown in Table~\ref{t10}. The experiments are conducted with a fixed 5-fold cross-validation scheme across all models. The model performance is further checked on Kaggle leaderboard of the ISIC20 dataset, where it achieved a single model AUC score of 0.9450 on unseen test data. It is mention-worthy that 

\subsection{Ablation Study}

\subsubsection{Effect of different backbone as feature extractor}

The baseline experiments are conducted using conventional CNN models, where EfficientNets\cite{efficientnet} and DenseNets\cite{densenet} outperformed other models. The ablation studies on the highest-performing models are detailed in Table~\ref{t1}. From the table, it can be observed that the proposed modules perform well across architectures with varying depths. The highest performance is observed on EfficientNetB5, which our proposed models augment even further.

\subsubsection{Effect of the contextual feature}
Only a single image (no contextual images) is used in the baseline model which mimics the traditional deep learning-based approaches. The performance of different modules with and without contextual images is provided in Table~\ref{t1}. The inclusion of only Contextual Feature (CF) and Comparative Contextual Feature(CCF) results in improvements of 2.8\% and 3.2\%, respectively, while including both provides an improvement of 4.5\%  for the VGG16 backbone. Furthermore, the inclusion of MKSA modules with only CFF and CCFF modules provides around 3.9\% and 3.7\% improvements while including the MKSA module with both CFF and CCFF modules results in 5\% improvement over baseline with VGG16 backbone. Hence, contextual features (CFF \& CCFF modules) are contributing to considerable improvements over the baseline performance. Similar improvements can also be noticeable for other backbones as well.

\subsubsection{Effect of the Multi Kernel Self Attention (MKSA) Module}

In Table~\ref{t1}, at the first stage (without contextual information), the introduction of this block to the backbone improved the AUC score by 1.175\% on average across the architectures used. When used in conjunction with the CCF block, this block increased the AUC score too, which can be observed from the table. The table also demonstrates that this module boosted the AUC score when used with the entirety of the proposed solutions.

%\subsubsection{Effect of different Contextual Feature Fusion Schemes}

%\subsubsection{Effect of the differential feature}

\subsubsection{Effect of the number of contextual images}
In practice, it is often not possible to have exactly 10 contextual images for each patient, and Clinicians are often required to base their decisions on the images available, which would vary from patient to patient. To adapt to this case, the model is required to be able to make a decision based on a random number of images per patient. 

To model this unique problem in the training method, a new augmentation is introduced. In this method, contextual images are randomly replaced with images that are black throughout, and their corresponding labels are set to 0. The model is then trained on these images as input, where the model is optimized to perform even when there is no relevant information in some of the input images. This is done to make the model more robust to the availability of contextual images. The model performance is found to be robust with changing the number of available contextual images. Even with up to 50\% of the contextual images blacked out randomly, the AUC score is 0.891, and with only 1 image randomly blacked out, the score increases to 0.90.

\subsubsection{Effect of the proposed Multi-phase optimization algorithm}

Initially, the module consisting of the baseline feature extractor with an MKSA block is trained on the dataset without any contextual module with a learning rate of $\gamma_1$ = 0.01, which scored slightly better than the baseline performance. Then the same module weights are used to initialize the primary and contextual feature extractors that are used with the subsequent modules. In the second phase of the experiment, the feature extractor weights are frozen and only the CFF and CCFF submodules are trained with a learning rate of $\gamma_2$ = 0.0075. This boosted the score by 2\% on average. And in the final phase, the feature extractor weights are unfrozen and trained with a learning rate of $\gamma_3$ = 0.006. On average, this phase yielded another 1.8\% boost.

\subsection{Qualitative analysis of the effect of contextual information}

To analyze the improvement of the proposed solution over the baseline, statistical results such as the AUC score have been used as the determining metric. The effect of contextual information on the model's decision is summarized in Fig.~\ref{pvsc}. As this problem is a ranking prediction problem, a sample having more confidence means that sample is more likely to be Melanoma. In Fig.~\ref{pvsc} each row is associated with one patient where the first column indicates the primary image and the rest of the column indicates contextual images. On each image there are $3$ labels; on the top left, ground truth is shown, and on the top right the model prediction is shown with (blue) and without (black) contextual images respectively, separated by a vertical line. From Fig.~\ref{pvsc} it can be conceived that identifying Melanoma using only primary images is very difficult. Hence, a model without contextual images fails to provide a confident and accurate prediction. For primary images (1st column) it is apparent that the proposed method provides a confident and accurate prediction for Melanoma whereas the traditional method provides a doubtful and less accurate prediction. As the proposed method utilizes comparative features along with contextual features, it has been able to provide a more improved decision upon observing the patient skin condition from the multiple contextual images hence resulting in a more confident and accurate state.

\section{Conclusion}
In this paper, an automated scheme is proposed to incorporate patient-level contextual information for accurate diagnosis of Melanoma diseases that provides exceptional performances with substantial improvement of AUC score over the existing approaches. The proposed method exploits single image-based features along with contextual and comparative features to assess Melanoma. It computes contextual features using a multi-kernel-based self attention mechanism and comparative features using additional pairwise comparison. The proposed scheme to identify Melanoma imitates clinicians' diagnosis strategy which can be employed for a clinical explanation. The experimental results indicate that introducing this scheme contributed to increased performance. Therefore, the proposed scheme can be easily employed for identifying Melanoma that can be an effective alternative to other state-of-the-art approaches.

\bibliographystyle{IEEEtran}
\bibliography{REFERENCE}

\end{document}